\documentclass[letterpaper]{article} 
\usepackage[]{aaai25}  
\usepackage{times}  
\usepackage{helvet}  
\usepackage{courier}  
\usepackage[hyphens]{url}  
\usepackage{graphicx} 
\usepackage{natbib}  
\usepackage{caption} 
\usepackage[english]{babel} 

\usepackage{algorithm}
\usepackage{algorithmic}
\usepackage{multicol}

\usepackage{newfloat}
\usepackage{listings}
\DeclareCaptionStyle{ruled}{labelfont=normalfont,labelsep=colon,strut=off} 
\floatstyle{ruled}
\newfloat{listing}{tb}{lst}{}
\floatname{listing}{Listing}
\usepackage[utf8]{inputenc} 
\usepackage[T1]{fontenc}    
\usepackage{booktabs}
\usepackage{enumitem}
\usepackage{tabularx}
\usepackage{fancyvrb}
\usepackage{fvextra}
\usepackage{multirow}
\usepackage{changepage}
\usepackage{threeparttable}
\usepackage{amsmath}
\usepackage{pgfplots}
\usepackage{pgfplotstable}
\usepgfplotslibrary{fillbetween}
\pgfplotsset{compat=1.16}
\usepgfplotslibrary{statistics}
\usepgfplotslibrary{groupplots}
\usepackage{caption}
\DeclareCaptionFont{helv}{\sf\small}

\usepackage{silence}
\pgfplotsset{every axis/.append style={
                    xlabel={$x$},          
                    ylabel={$y$},          
                    label style={font=\sffamily},
                    tick label style={font=\sffamily\scriptsize},
                    xticklabel style = {font=\sffamily\scriptsize},
                    title style = {font=\normalsize\sffamily},
                    ylabel near ticks,
                    y label style={font=\sffamily\scriptsize},
                    xlabel near ticks,
                    x label style={font=\sffamily\scriptsize},
                    legend cell align={left},
                    legend style={draw=none, font=\sffamily\scriptsize},  
                    },
                    }

\newcommand*{\eg}{{\em e.g.}}

\newcommand*{\ie}{{\em i.e.}}

\title{Citations and Trust in LLM Generated Answers}
\author{
    Yifan Ding\textsuperscript{\rm 1}, Matthew Facciani\textsuperscript{\rm 1}, Amrit Poudel\textsuperscript{\rm 1}, Ellen Joyce\textsuperscript{\rm 1}, Salvador Aguinaga\textsuperscript{\rm 2}, Balaji Veeramani\textsuperscript{\rm 2}, Sanmitra Bhattacharya\textsuperscript{\rm 2}, Tim Weninger\textsuperscript{\rm 1}\thanks{This study is being funded by the United State Defense Advanced Research Projects Agency under contracts HR001121C0168 and HR00112290106.}
}
\affiliations{
    \textsuperscript{\rm 1}Department of Computer Science and Engineering, University of Notre Dame, Notre Dame, IN 46556\\
    \textsuperscript{\rm 2}AI Center for Excellence, Deloitte \& Touche LLP, New York City, NY, 10112\\
    \{yding4, mfaccian, apoudel, ejoyce3, tweninger\}@nd.edu,  
    \{saguinaga, bveeramani, sanmbhattachary\}@deloitte.com \\
}

\begin{document}
\nocopyright 
\maketitle

\begin{abstract}
Question answering systems are rapidly advancing, but their opaque nature may impact user trust. We explored trust through an anti-monitoring framework, where trust is predicted to be correlated with presence of citations and inversely related to checking citations. We tested this hypothesis with a live question-answering experiment that presented text responses generated using a commercial Chatbot along with varying citations (zero, one, or five), both relevant and random, and recorded if participants checked the citations and their self-reported trust in the generated responses. We found a significant increase in trust when citations were present, a result that held true even when the citations were random; we also found a significant decrease in trust when participants checked the citations. These results highlight the importance of citations in enhancing trust in AI-generated content.
\end{abstract}

%

\section*{Introduction}
Large language models (LLMs)~\cite{achiam2023gpt4, touvron2023llama2} stand at the forefront of contemporary artificial intelligence (AI), wielding immense potential to reshape the way humans interact with technology and information. However, a critical question persists: Are these LLMs and their by-products trusted by users? The answer holds profound implications for the future of AI in society. Trust~\cite{baier1986trust}, a cornerstone of human relationships and societal functioning, plays an indispensable role in the sharing and acceptance of information.

Human trust dynamics are complex, often deeply rooted in social norms and interpersonal relationships. Humans navigate complex social structures by building trust through shared experiences, reputations, and accountability mechanisms~\cite{muir1987trust}. Leading social theories, like the Principle of Social Proof, suggest that social conventions might predispose individuals to favor human sources over algorithmic  sources~\cite{cialdini2009social}. This predicts that responses from an AI system might be more trusted when a human source corroborates its response. Conversely, responses from an AI system might be more trusted when the human touch is absent because AI systems lack anthropomorphic traits and social nuances~\cite{miller2019explanation}. In other words, AI systems, which are largely devoid of social motives, may increase trust among users who seek impartial and objective information~\cite{sambrook2012delivering,rosen1999journalists}.

Human trust in AI is a complicated subject that depends on several factors, with \textit{accuracy} being the most important contributor~\cite{lucassen2011factual} and \textit{explainability}~\cite{rawal2021recent} following not far behind. Accuracy refers to the AI's ability to provide correct information consistently. Explainability entails the articulation of the model's decision-making pathways, rendering them transparent and comprehensible to users. Explainability is crucial for building a trust-based relationship between humans and AI~\cite{stephanidis2019seven}, significantly influencing users' willingness to engage with AI systems~\cite{hoff2015trust}.
 
\begin{figure}[t]
    \centering
    \includegraphics[width=0.75\linewidth]{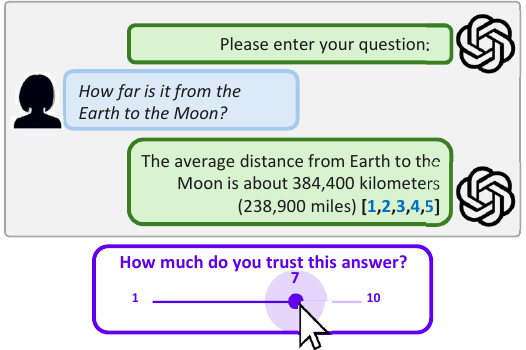}
    \caption{AI Chatbot system answering a user's question with five hyperlink citations. The presence of citations significantly increases the user's trust of the response.}
    \label{fig:teaser}
\end{figure}

Existing research on explainability primarily focuses on enabling AI engineers to understand model behavior. However, empirical investigations of explainability's role in user trust are limited, and the evidence is mixed regarding whether explainability indeed increases user trust~\cite{scharowski2023exploring,nothdurft2013impact}. Some studies report positive effects~\cite{ehsan2019automated}, while others find no significant impact~\cite{poursabzi2021manipulating,zhang2020effect,cheng2019explaining}. These mixed results may stem from differences in how user trust is defined and measured: some studies rely on simple questionnaires, like in Fig.~\ref{fig:teaser}, while others assess user trust through behavior-related metrics~\cite{poursabzi2021manipulating}. This distinction is important because, in studies, an increase in user trust does not necessarily translate to increased adoption or reliance on the AI~\cite{papenmeier2022s,miller2016behavioral}. 

Because recent LLMs are exceedingly complex, the focus of explainability in AI has shifted from model interpretability---designing models so that their decision-making processes can be visualized or observed~\cite{azaria2023internal}---to \textit{post-hoc explanations}, which denote how a specific decision was reached after it has been made~\cite{lipton2018mythos}. These post-hoc explanations typically take the form of feature importance and counterfactual explanations~\cite{wachter2017counterfactual}. Feature importance explanations identify which variables most influenced a particular outcome, helping users understand the model's reasoning process~\cite{zhao2024explainability}. Counterfactual explanations, on the other hand, describe how the model's output would change if certain inputs were different, providing insights into the decision-making logic~\cite{mothilal2020explaining}.

In LLMs systems, explainability is often conveyed via citations, which represent tangible evidence of the source of a system's knowledge and lends credibility to the response. By properly citing sources, these systems acknowledge the origins of their ideas and support their arguments with evidence. This practice not only respects the intellectual property of others but also enhances the overall quality and credibility of the response~\cite{thornley2015role}. 

The widespread adoption of LLMs has led to the development of Retrieval Augmented Generation (RAG) systems. These systems generate explanations by incorporating external information retrieved from various sources~\cite{lewis2020retrieval}. By extracting and integrating relevant external data, RAG systems oftentimes provide explicit citations in order to improve the transparency and reliability of their output and to encourage users to further explore the topic~\cite{srinivas2024perplexity}.

However, the role that citations play in shaping the trust relationship between AI and their users is not well understood. In the present study, we describe the results of a set of experiments that asks:

\begin{enumerate}
    \item Do citations increase self-reported user trust in LLM-generated responses? If so, does the number of citations matter? and does the relevance of the citations matter? 
    \item Does the act of checking the citation decrease self-reported user trust?
\end{enumerate}

Our study applies the social theories of \textit{trust as anti-monitoring} and \textit{the principle of social proof} to contextualize factors influencing user trust in LLMs. We hypothesized that (1) providing users with the sources behind LLM responses via citations (\ie, social proof) would enhance trust in these otherwise opaque systems and that (2) providing users with the ability to check the citation would increase user trust; however, the act of actually checking citation (\ie, monitoring the LLM) would be an indication of a lower level of user trust in the LLM's response. 

To test these hypotheses, we deployed a bespoke QA Web site and invited participants to submit open-ended questions. LLM-generated responses were returned to the user with varying numbers of citations (zero, one, or five), which were either relevant to the answer or randomly selected from previous queries. Our analysis revealed a statistically significant increase in self-reported user trust when citations were included, a result that held true even when the citations were random. We also found a statistically significant decrease in self-reported user trust when participants checked the citations.

These findings underscore the critical role of citations in bolstering user trust in LLM-generated content. Moreover, they illuminate the nuanced interplay between citation relevance, question context, and user perceptions of trust.

\subsection*{Trust and Large Language Models}

Trust in AI is a flourishing research area with diverse perspectives, including Trust as Anti-Monitoring and Social Proof Theory integrated into RAG systems and Chatbots. Scholars across these domains have investigated the dynamics of user trust with AI, offering insights into the mechanisms shaping human-machine interactions.

The first step when investigating users' trust in LLMs is to define trust. Trust is context-dependent, for example, a person may trust a mechanic to repair their car but not to prepare their taxes.  Additionally, trust is dynamic and is built over a series of human interactions or events; each time the mechanic successfully repairs the car, the individual's trust increases, whereas a failure to fix it properly would decrease their trust.

In scenarios such as judicial decisions and wikis, citations are crucial in building trust, as they add credibility and transparency to content, regardless of the category of content. This raises an intriguing question: do individuals inherently trust LLM-generated responses more when they are accompanied by citations? 

\begin{figure*}
    \centering
    \includegraphics[width=1.0\linewidth]{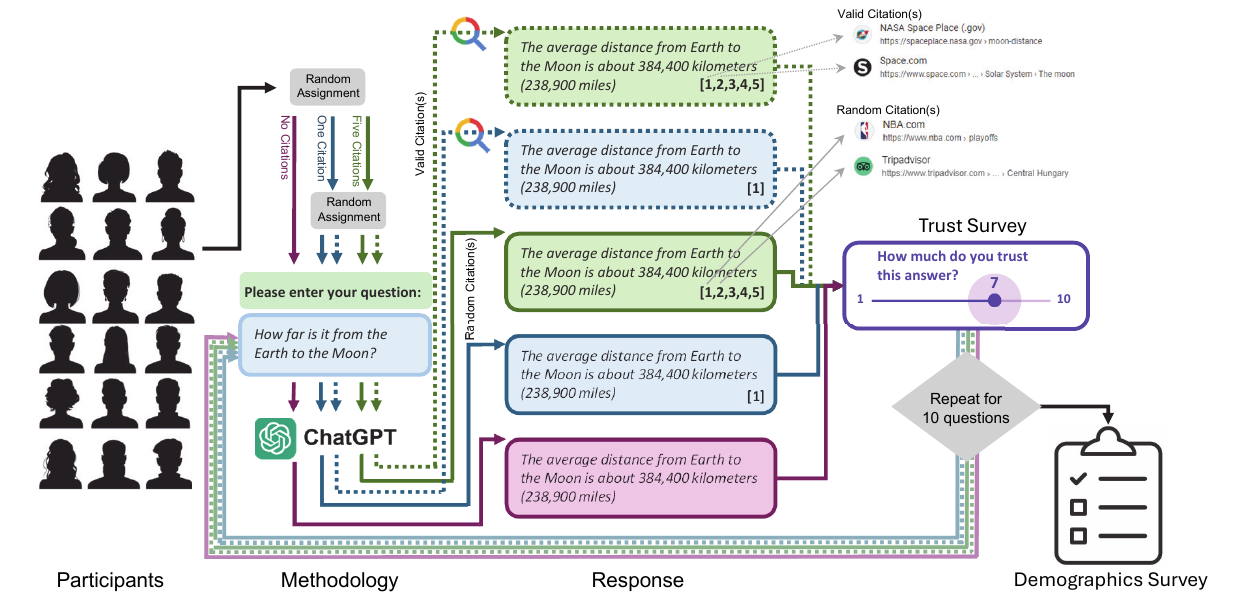}
    \caption{Methodology of the Citation Trust Experiment. Participants are assigned to zero (\textcolor{purple}{purple}), one (\textcolor{blue}{blue}), or five (\textcolor{green}{green}) citations, which can be either valid (dotted-line) or random (solid-line). A participant may ask any question, and then rates the response on a scale of 1 to 10. This is repeated for ten total questions and a demographics survey is asked at the end.}
    \label{fig:framework}
\end{figure*}

\paragraph{Trust as Anti-Monitoring} How best to measure trust is a complicated and hotly debated topic~\cite{baier1986trust,ferrario_how_2022}. Trust, as Annette Baier's work suggests, can be construed as ``anti-monitoring,'' where an indication of trusting an entity is a reduction in monitoring their behavior~\cite{baier1986trust,archard2013reading}. Monitoring, in the example above, refers to the intuition that if a customer trusts a mechanic, the customer is willing to allow the mechanic to repair the car without supervision. This implies that the level of monitoring someone performs is inversely related to the level of trust they have in the person or thing being monitored. The concept of trust as anti-monitoring offers a measurable framework for understanding trust. Citations serve as a monitoring mechanism, allowing people to check the LLM's response and determine if it aligns with the user's expectations and reasoning. This ties into the social-psychological Principle of Social Proof, where individuals look to external cues and validations to form trust.

\paragraph*{Principle of Social Proof} 

The Principle of Social Proof is particularly useful for understanding interactions with Chatbots. This framework suggests that people are more likely to adopt a behavior if they see the social proof of others doing the same~\cite{cialdini2009social}. Social proof can, therefore, act as a proxy for trustworthiness, as individuals are more likely to use, and in turn trust, a product when they observe others using it~\cite{lins2023advancing,kim2024medical,venkatesh2000theoretical}.

In Chatbot interactions, citations within the outputs serve as strong indicators of social proof because they signal to users that the output is endorsed by some source. Consequently, the presence of citations theoretically enhances trust, because users perceive the content as more credible and reliable.

The Principle of Social Proof aligns with the anti-monitoring framework of trust, where having the ability to monitor or check a response positively contributes to the trustworthiness of a response, regardless of whether or not the user chooses to check the source. However, it remains uncertain whether high-quality citations significantly impact trust, or if \textit{any citation}, regardless of quality, is sufficient.


\paragraph*{RAG and Chatbots} RAG is an AI framework that incorporates information retrieved from external sources into the generation process \cite{asai2023self,asai2024reliable}. This approach tends to reduce inaccurate responses\cite{lewis2020retrieval} and includes citations within its generated responses, providing users with a clear trail of the sources that support the presented information. 

User trust dynamics in LLMs is beginning to receive some much-needed attention and research has uncovered factors that influence users' perceptions and behaviors \cite{sun2022recitation}. For example, users are more likely to engage with chatbots they perceive as trustworthy \cite{choudhury2023investigating}. However, concerns about government use of chatbots can lead to distrust \cite{aoki2020experimental}. Despite occasional inaccuracy and unreliability in current chatbot versions, users often express intentions to continue using them, indicating a resilient trust in these systems \cite{amaro2023ai}. Research also suggests that users tend to trust chatbots with more human-like characteristics \cite{kaplan2023trust}, highlighting the interplay between trust, utility, perceived reliability and accuracy, and the humanization of AI in shaping user engagement.

\section{Citations and Trust Experiment}

We developed a bespoke Web site for data collection. On this Web site, users were introduced to the task with an animation that showed the example question: ``How far is it from the earth to the moon?''. If the participants agreed to participate, they were provided a simple query box, stylized to look like a standard input Web form. On this form users were prompted by instruction-text to: ``Ask any question''. 

The participant's responses were stored on a Web server owned and managed by research team. Each question was then fed directly to ChatGPT4 and the responses were collected. 
Responses were truncated if they were longer than three sentences. 

The experiment was a Randomized Controlled Trial (RCT) with a between-subjects 3 by 2 factorial design (see Fig.~\ref{fig:framework}). The first factor corresponded to the number of citations: zero, one, or five; the second factor corresponded to the nature of the citations: valid or random. 

For the first factor: in the no citation condition, the response was taken directly from the output of ChatGPT, truncated to three sentences if necessary, and provided to the participant. In the non-zero citation conditions, we redirected the truncated response from ChatGPT to a Web search API\footnote{\url{https://scaleserp.com}}, which queried the Google search engine for Web sites relevant to the response; the first five search engine results were recorded. This was invisible to the participant; however, there was a small response delay in this condition. In the one-citation condition, the top citation was provided to participant as a numeral (\eg, [1]). In the five-citation condition, all five citations were provided to the participant as a list of numerals (\eg, [1,2,3,4,5]). Each citation was programmed to show the URL of the citation if the participant hovered their mouse over the numeral (see Supplement C Figures S1 and S2).

For the second factor: in the valid citation condition, the search engine result(s) were provided directly to the user. In the random citation condition, the actual citations were recorded, but the citation URL(s) shown to the participant were randomly selected from citations of previous participant's questions. 

\subsection*{Participants}

We used Prolific\footnote{prolific.com} to recruit participants for this study~\cite{palan2018prolific}. Data collection occurred on March 13, 2024. Participants were paid two US dollars and took a median of 17 minutes to complete. The study had 303 total participants who were randomly assigned to the experimental groups (\ie, between-subjects design). Participants either saw zero (N=108), one (N=96), or five (N=101) citations. Of the two groups who saw citations (N=197), a random split of the participants received a random citation (N=87) or valid citation (N=110). 

Participants were asked to enter ten questions and rate each response; finally an exit interview was conducted with a battery of demographic questions (see Supplement A and Table S1 for demographic battery and response codes). 

\subsection*{Ethical Statement}

This study received approval from the \textit{redacted} Institutional Review Board (protocol no. \textit{redacted}
). Participants were fully briefed on the study's purpose, ensuring informed consent and voluntary participation. Aside from broad demographic questions, personal identifiable information was not collected. 

Our study aims to understand and measure trust and inform best practices in the incorporation of citations into AI systems. However, potential risks include misinterpretation of results leading to over-reliance on citations and privacy concerns related to participant data. We are committed to addressing these issues by adhering to ethical standards, ensuring transparency, and carefully evaluating the broader impact of our work.




\subsection*{Data, Materials, and Software Availability}
All participant questions, their responses and, their ratings are available in an Excel file. This file is publicly available online at \textit{redacted} and in the Supplement. 
We used Stata Software for data and statistical analysis. The Stata codes used for all analysis in the main manuscript is included in the Supplementary Material.

\section*{Results}

\subsection*{Do Citations Increase User Trust?}

\begin{figure}[t]
    \centering
    \begin{tikzpicture}
    \begin{groupplot}[
        group style={group size=1 by 3,
            horizontal sep = .2 cm, 
            vertical sep = 0 cm,            
            ylabels at=edge left,
            xlabels at=edge bottom,
            xticklabels at=edge bottom,
            yticklabels at=edge left}, 
    width=.9\linewidth,
    xlabel={Trust},
    ylabel={},
    ylabel absolute, ylabel style={yshift=4em, font=\sffamily\scriptsize},
    xmax = 1,
    xmin = -1,
    ]
    \nextgroupplot[
    ylabel={},
    ymin = 2.25,
    ymax = 9.75,
    height=2.05in,
    ytick={3,4,5,6,7,8,9},
    yticklabels={ Age Group, Heard of ChatGPT, Education Level, Male, Conservative, Non-White, Urban},
    scatter/classes={a={blue}, b={red}, c={green}, d={black}},
    scatter,
    only marks,
    clip=false,
    ]%
    \addplot[dashed, samples=50, smooth, no marks, domain=0:6, black] coordinates {(0,3)(0,9.5)};
    \node[anchor=south east, font=\footnotesize\sffamily] at (axis cs:1,9.6) {*** $p$<0.01, ** $p$<0.05, * $p$<0.1};
    
    \addplot [
        scatter src=explicit symbolic,
    visualization depends on={value \thisrow{lab} \as \Label},
    nodes near coords*={\tiny{\sffamily{\Label}}},
    error bars/.cd, x dir = both, x explicit, error bar style={very thick ,solid, black},
      error mark options={line width=0pt,mark size=0pt,rotate=90}]
    table[meta=class, x=x, y=y, x error=ex]{
        y x     ex      class   lab        
        3 -0.017 0.0335 d {}
        4 0.112 0.1240 d  {}
        5 -0.041 0.0511 d {}
        6 0.082 0.0906 d {}
        7 -0.061 0.0386 d  {}
        8 0.232 0.0940 d **
        9 -0.132 0.0698 d * 
    };

    \nextgroupplot[
    ylabel={},
    ytick={1,2},
    ymin = 0.25,
    ymax = 1.75,
    height=0.9in,
    yticklabels={Has Citation},
    scatter/classes={a={blue}, b={red}, c={green}, d={black}, e={cyan}},
    scatter,
    only marks,
    ]%
    \addplot [
        scatter src=explicit symbolic,
    nodes near coords*={\tiny{\sffamily{\Label}}},
    visualization depends on={value \thisrow{label} \as \Label},
    error bars/.cd, x dir = both, x explicit, error bar style={very thick ,solid, cyan},
      error mark options={line width=0pt,mark size=0pt,rotate=90}]
    table[meta=class, x=x, y=y, x error=ex]{
        y x     ex      class   label        
        1 0.394 0.0906   e      ***
    };

    \nextgroupplot[
    ylabel={},
    ytick={1,2},
    ymin = 0.25,
    ymax = 1.75,
    height=0.9in,
    yticklabels={Random Citation},
    scatter/classes={a={blue}, b={red}, c={green}, d={black}},
    scatter,
    only marks,
    ]%
    \addplot [
        scatter src=explicit symbolic,
    nodes near coords*={\tiny{\sffamily{\Label}}},
    visualization depends on={value \thisrow{label} \as \Label},
    error bars/.cd, x dir = both, x explicit, error bar style={very thick ,solid, blue},
      error mark options={line width=0pt,mark size=0pt,rotate=90}]
    table[meta=class, x=x, y=y, x error=ex]{
        y x     ex      class   label        
        1 -0.268 0.0872 a   ***
    };

\end{groupplot}
\end{tikzpicture}
     \vspace{-0.7cm}
    \caption{Citations increase perceived trustworthiness, but random citations decrease perceived trustworthiness. Regression coefficients $\beta$ and their standard errors are plot on the x-axis.}
    \label{fig:citation_lr}
\end{figure}
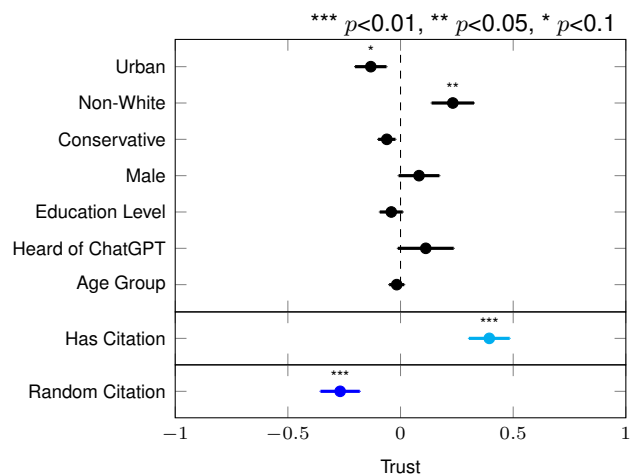

We expect that the presence of citations in an AI chatbot's output should enhance response transparency and should improve perceived trustworthiness. The Principle of Social Proof suggests that observing evidence of others endorsing a behavior increases the likelihood of adopting that behavior ourselves. We also predict that the quality of the citations matters. Citations that accurately support the chatbot's answer will be evaluated as more trustworthy than random citations. 

In out initial analysis, the Dependent Variables (DVs) were citation and no citation coded 1 or 0 respectively. Random and Accurate citation were coded 1 or 0 respectively. Controlling for various demographic factors, results of linear regression analyses indicate a statistically significant increase in perceived trustworthiness for AI chatbot responses with citations compared to those without (See Fig.~\ref{fig:citation_lr} and Supplement B Table S2).

\subsection*{Does the Quality of Citation Matter?}
Additional analysis examined the influence of random versus valid citations on trustworthiness. Again, controlling for demographic factors, results revealed that answers containing random citations were significantly less trustworthy (See Fig.~\ref{fig:citation_lr} and Supplement B Table S2).

\subsection*{Do the Number of Citations Matter?}
Using an analysis of variance (ANOVA), we examined whether perceived trustworthiness varied among the zero citation, one citation, and five citation conditions. The analysis revealed significant differences between groups ($F(2, 3037) = 10.23$, $p<.001$). Post-hoc Bonferroni~\cite{bonferroni1936teoria} tests indicated that both the one citation and five citation conditions were rated significantly higher on trustworthiness compared to the zero citation condition (see Supplement B Table S3). 

Notably, there was no significant difference between the one citation and five citation conditions ($p>.05$). In other words, five citations in an answer are not perceived as more trustworthy than an answer with one citation. This negative result is contrary to our initial hypothesis. One plausible reason for this result might be due to the principle of diminishing returns, wherein participants may perceive that a single, well-chosen citation is sufficient to confirm the AI's response. 

\subsection{Social Demographics and Trust}

In addition to the directional hypotheses, we also explored how various social demographics predict the perceived trustworthiness of AI chatbot answers. While there is limited research specifically on how different groups respond to AI chatbots, some studies investigate how different demographics react to new technologies such as AI~\cite{tyson2023growing,rainie2022americans}. For instance, individuals with more conservative values tend to be more skeptical toward AI and new technologies~\cite{castelo2021conservatism}. However, both individuals with liberal and conservative views are more receptive to technology when it is framed in a way that aligns with their political values~\cite{claudy2024should}. Furthermore, higher levels of education and familiarity with chatbots may increase perceived trustworthiness; on the other hand, greater awareness of AI limitations, which often accompany higher education and familiarity, could decrease trustworthiness.

We did not find significant differences in trustworthiness ratings among most demographic categories, the slight inclination of nonwhite participants to trust the answers more suggests avenues for exploratory research (SI Appendix B Table S2). For instance, future studies could look further into the underlying factors driving this trend and explore potential cultural or societal influences on trust perceptions in AI-generated content.

\begin{figure}[t]
    \centering
    \begin{tikzpicture}
    \begin{groupplot}[
        group style={group size=1 by 3,
            horizontal sep = .2 cm, 
            vertical sep = 0 cm,            
            ylabels at=edge left,
            xlabels at=edge bottom,
            xticklabels at=edge bottom,
            yticklabels at=edge left}, 
    width=0.9\linewidth,
    xlabel={Trust},
    ylabel={},
    ylabel absolute, ylabel style={yshift=4em, font=\sffamily\scriptsize},
    xmax = 1,
    xmin = -1,
    ]

    \nextgroupplot[
    ylabel={},
    ymin = 0.25,
    ymax = 9.75,
    height=2.57in,
    ytick={1,2,3,4,5,6,7,8,9},
    yticklabels={Random Citation, Has Citation, Age Group, Heard of ChatGPT, Education Level, Male, Conservative, Non-White, Urban},
    scatter/classes={a={blue}, b={red}, c={green}, d={black}},
    scatter,
    only marks,
    clip=false,
    ]%

    \node[anchor=south east, font=\footnotesize\sffamily] at (axis cs:1,9.6) {*** $p$<0.01, ** $p$<0.05, * $p$<0.1};
    
    \addplot[dashed, no marks, samples=50, smooth,domain=0:6, black] coordinates {(0,0)(0,9.8)};

    \addplot [
        scatter src=explicit symbolic,
    visualization depends on={value \thisrow{lab} \as \Label},
    nodes near coords*={\tiny{\sffamily{\Label}}},
    error bars/.cd, x dir = both, x explicit, error bar style={very thick ,solid, black},
      error mark options={line width=0pt,mark size=0pt,rotate=90}]
    table[meta=class, x=x, y=y, x error=ex]{
        y x     ex      class   lab        
        1 -0.211    0.0490 d    ***
        2 .423  0.0510 d ***
        3 0.005 0.0118 d {}
        4 0.058 0.0696 d  {}
        5 -0.029 0.0287 d {}
        6 0.202 0.0509 d ***
        7 -0.086 0.0217 d  ***
        8 -0.072 0.0528 d {}
        9 0.153 0.0392 d *** 
    };

    \nextgroupplot[
    ylabel={},
    ytick={1},
    ymin = 0.25,
    ymax = 1.75,
    height=0.90in,
    yticklabels={Check Citation},
    scatter/classes={a={blue}, b={teal}, c={green}, d={black}},
    scatter,
    only marks,
    ]%
    \addplot [
        scatter src=explicit symbolic,
    nodes near coords*={\tiny{\sffamily{\Label}}},
    visualization depends on={value \thisrow{label} \as \Label},
    error bars/.cd, x dir = both, x explicit, error bar style={very thick ,solid, teal},
      error mark options={line width=0pt,mark size=0pt,rotate=90}]
    table[meta=class, x=x, y=y, x error=ex]{
        y x     ex      class   label        
        1 -0.058 0.0104 b ***
    };

    \addplot[dashed, samples=50, smooth,domain=0:6, black] coordinates {(0,-1)(0,10)};

    \nextgroupplot[
    ylabel={},
    ytick={1,2,3},
    ymin = 0.25,
    ymax = 3.75,
    height=1.25in,
    yticklabels={Political, Factual, Complex},    
    scatter/classes={a={blue}, b={red}, c={orange}, d={black}},
    scatter,
    only marks,
    ]%
    \addplot [
        scatter src=explicit symbolic,
    nodes near coords*={\tiny{\sffamily{\Label}}},
    visualization depends on={value \thisrow{label} \as \Label},
    error bars/.cd, x dir = both, x explicit, error bar style={very thick ,solid, orange},
      error mark options={line width=0pt,mark size=0pt,rotate=90}]
    table[meta=class, x=x, y=y, x error=ex]{
        y x     ex      class   label                
        1 0.314 0.156   c      **
        2 0.374 0.0865   c      **
        3 0.0697 0.275   c      {}
    };

    \addplot[dashed, samples=50, smooth,domain=0:6, black] coordinates {(0,-1)(0,10)};

\end{groupplot}
\end{tikzpicture}

    \vspace{-0.7cm}
    \caption{Checking citations decrease perceived trust. Political and factual questions have a higher perceived trust. Regression coefficients $\beta$ and their standard errors are plot on the x-axis.}
    \label{fig:rq12}
\end{figure}
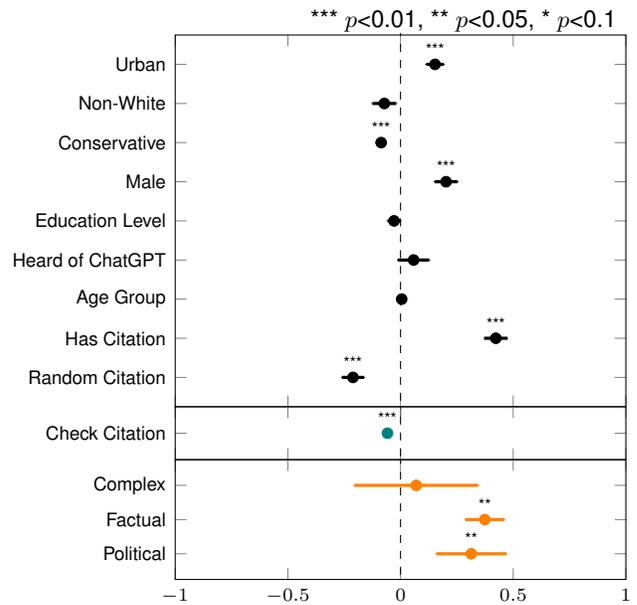

\subsection*{Does Checking Citations Indicate a Reduction in User Trust?}
The theory of trust as anti-monitoring predicts that individuals who are skeptical of an answer will be more likely to check the source of the citation. This predicts that checking a citation indicates reduced trust, as users are no longer relinquishing control and trusting the other party to be accurate. We tracked the frequency of participants that manually checked citations with their mouse while reviewing the AI chatbot's answers. There were 1,976 answers in our dataset that had at least one reference. Of these, only 193 (9.77\%) were manually checked. 83 participants out of 197 participants (42.1\%) in the citation groups checked at least one citation while completing the experiment. Interestingly, participants in the five citation group were significantly more likely to do a citation check ($\chi^2$= 21.19; $p$<0.001). 

We coded Check Citation as 0 or 1 and controlled for all demographic variables, as well as the presence of citations and random citations. Trust was included as our main independent variable (IV). Linear regression, controlling for demographics as well as the presence of citations and random citations, was performed. The analysis (see Fig.~\ref{fig:rq12} and Supplement C Table S4) revealed a significant correlation between increased citation checks and lower trust ratings for the answers.  These findings lend support to the trust as anti-monitoring theory: a higher frequency of citation checking predicts lower trustworthiness.

These results raise the broader question of whether \textit{any} citations, even if they were random, would yield higher trust ratings than zero citations. We compared the no citation trust ratings with the trust ratings of random citations for questions that were checked. We did not find a significant difference between answers that had zero citations and answers that had random citations that were checked (T=-0.877; $p$=0.38). There were only 193 checked-questions to analyze, but we did see a lower mean trust rating in the checked random citation answers (M=7.55; SD=2.54) compared to the no citation answers (M=7.73; SD=2.46) (see SI Appendix C Table S5). In other words, answers with random citations that were manually checked were no more trusted than answers with no citations at all.

We also found significant differences based on gender, political orientation, and residence. Specifically, males, those indicating liberal political orientation, and individuals residing in urban areas were significantly more likely to check the citations ($p < .001$) (see SI Appendix C Table S4).

\subsection{Illustrating Question Semantics}
The data we collected includes hundreds of interesting and unique real-world, human-generated questions, along with trust ratings for their answers. 

This data permitted an exploration of the types of questions asked and if they affected trustworthiness. Political information is particularly susceptible to bias~\cite{ditto2019least, poudel2024navigating}, and we believe that questions of a political nature may vary in their perceived trustworthiness. Individuals might be more likely to ask questions that confirm their pre-existing political biases, which could increase the perceived trustworthiness of the answer if it aligns with their views. Conversely, if the chatbot's answer contradicts their beliefs, the perceived trustworthiness of the response may decrease.

\begin{figure}[t]
    \centering
    \includegraphics[width=1\linewidth]{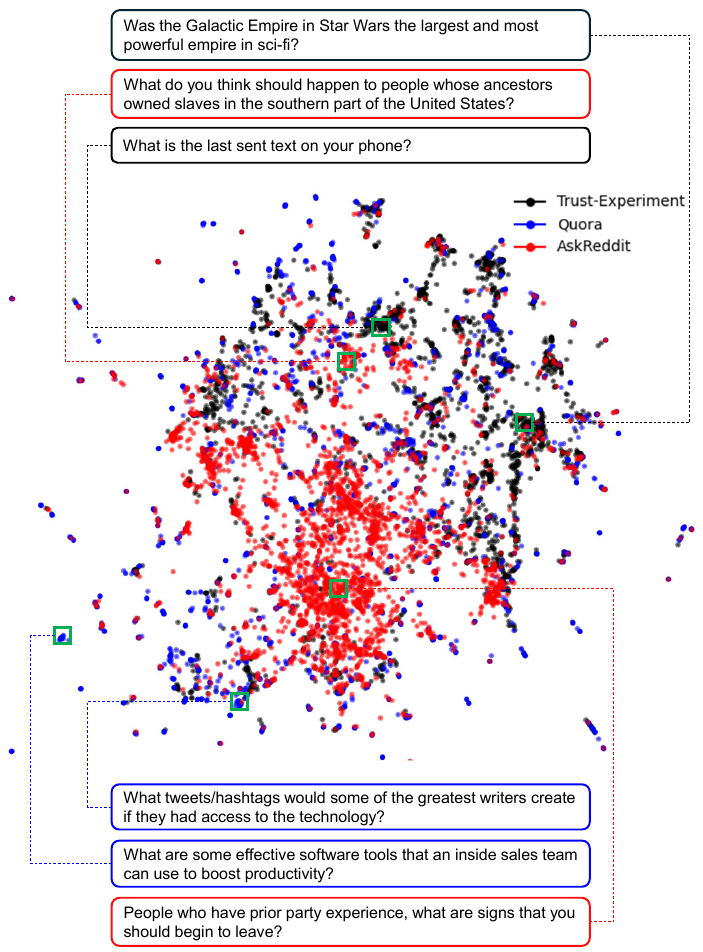}
    \caption{Visualization of question-topics asked by participants. Although we observe a substantial overlap in the kinds of questions asked by our participants (black) compared to AskReddit (\textcolor{red}{red}) and Quora (\textcolor{blue}{blue}), we also identify several topical gaps. Some representative samples of these topical gaps are illustrated in on right. An interactive visualization of this figure is included in the supplementary material.}
    \label{fig:rq4}
\end{figure}

We analyzed the semantics of the questions asked by participants using a comparative approach by juxtaposing them with questions from established question-answering datasets such as AskReddit and Quora. Using sentence transformers, we embedded each question into a high-dimensional vector space, capturing their semantic representations~\cite{reimers-2019-sentence-bert}. Subsequently, we used UMAP, a dimensionality reduction technique, to project these question embeddings onto a 2-dimensional plot~\cite{mcinnes2018umap}, as depicted in Fig.~\ref{fig:rq4}. In this plot, participant's questions are depicted in black. An interactive plot can be found in the Supplement.

\subsection*{Does the Type of Question Impact User Trust?}

Using linear regression, and accounting for demographics, citations, and random citations, we examined the impact of question type on user trust ratings. We then used ChatGPT4 to label each question as being Political, Factual, and Complex questions as 0 or 1; these labels need not be disjoint (\ie, a question can be both factual and political). 

Our results indicate that political questions received significantly higher self-reported user trust ratings compared to non-political questions\footnote{see Fig.\ref{fig:rq12}, Supplement D Table S6; the list of questions and their labels is also included in the Supplement}, even after adjusting for demographic factors. Manual analysis of the questions and answers suggests that this effect is not due to the chatbot's awareness of the participant's political stance, but rather because political questions are often framed in a way that aligns with users' preconceived notions, thereby eliciting more favorable responses. These findings support the theory of social proof, where the chatbot’s responses act as a form of social endorsement, enhancing perceived trustworthiness.


We found that fact-based questions were significantly more trusted than non-fact-based questions. This aligns with the previous result that political questions (which may also be factual) are also trusted, but it highlights different aspects of trustworthiness in AI chatbot responses. As previously discussed, political questions tend to be framed in a way that aligns with the participant's pre-existing beliefs and ingroup biases, leading to higher trust ratings. Furthermore, fact-based questions are typically grounded in objective, verifiable information. Participants can cross-check these facts against known data or their existing knowledge, leading to higher trust ratings. These findings suggest that trust in AI chatbot responses can be driven by both the objective accuracy of the information (fact-based questions) and the social alignment with the participant's beliefs (political questions). While fact-based questions benefit from their verifiability, political questions benefit from framing and ingroup validation.


\subsection{Complexity and Trust}
Previous studies suggest that LLMs provide more accurate responses when prompted with more familiar language~\cite{gonen2022demystifying}.  For example, asking ``Who was the first president of the United States?'' is clear and straightforward, leading to an accurate response. In contrast, a more complex question like ``How have the economic policies of U.S. presidents influenced income inequality in the United States?'' is less direct, causing the AI to infer more, which may reduce accuracy. This suggests that simpler, familiar prompts yield more accurate and trusted responses from AI systems.

Language perplexity was used as a proxy for the model's familiarity with the question, where perplexity was measured as $\text{PPL}(x) = \exp\left(-\frac{1}{N} \sum_{i=1}^{N} \log_2 P(w_i)\right)$ where $N$ is the total number of words in the response, $w_i$ is the $i^\textrm{th}$ word in the response, and $P(w_i)$ is the is the probability of the $i^\textrm{th}$ word given in ChatGPT4. In this context, lower perplexity indicates a more familiar prompt, leading to more accurate (perhaps more trusted) results. Guided by these prior findings, we compared user-reported trust as a function of the perplexity of the prompt. We found that higher perplexity is slightly (negatively) correlated with trust (Pearson R=-0.06, $p$=0.002), \ie, answers to complex questions are (slightly) less trusted.

In a similar exploratory analysis, we also found that the length of a prompt (number of tokens) is (slightly) negatively correlated with trust (Pearson R=-0.04, $p$=0.041). In other words, although we do not find statistical differences in trust between simple and complex questions, we did find that responses to simpler questions (in terms of perplexity and length) were more trusted. This suggests that users may have a higher level of trust in chatbots when the prompts are simpler, potentially indicating a preference for straightforward and concise queries that yield more understandable answers.

\section*{Discussion}

The present study investigated how variations in citations influenced the perceived trustworthiness of answers provided by an AI chatbot. Drawing upon the anti-monitoring framework, we conceptualized trust and extended this framework to incorporate the Principle of Social Proof. We hypothesized that participants would trust AI chatbot responses with citations more than those without citations, as citations provided the opportunity for verification (or monitoring) of the output. Moreover, these citations, linked to supporting organizations, served as a form of social proof, enhancing trustworthiness.

Our findings supported this hypothesis, revealing that AI chatbot answers with citations were perceived as more trustworthy compared to those without citations. Furthermore, we investigated the significance of the number of citations, finding no significant difference in perceived trustworthiness between responses with one or five citations.

Beyond the number of citations, we also investigated those participants who manually inspected the citations. We found that responses containing random citations were rated lower in trustworthiness compared to those with accurate citations.

Furthermore, we explored whether a higher frequency of citation-checks, indicated by mouse hovers, correlated with lower perceived trustworthiness. Consistently, participants who checked citations tended to rate the answers as less trustworthy. This finding aligns with the theory of trust as anti-monitoring, as skeptical participants sought to verify (\ie, monitor) the source of the information provided by the chatbot. Next, we investigated whether the type of question asked was associated with the perceived trustworthiness of the answers. We categorized questions into three groups: political, factual, and complex. 

Our analysis revealed that political questions were rated significantly more trustworthy than non-political questions. This trend may be attributed to participants posing political questions that already aligned with their political biases, leading them to be more inclined to trust the answers. We also found that fact-based questions were significantly more trusted than others. This difference could be due to the concrete nature of factual questions, instilling confidence in participants that they already know the correct answer, whereas non-factual questions were more subjective.

Regarding complex questions, we hypothesized that they might elicit greater trust as they could potentially demonstrate the chatbot's capability to handle challenging inquiries. However, we did not observe a significant difference in trust ratings between complex problem-solving questions compared to those categorized as more straightforward. Given the limited number of questions coded as complex, we cannot assert the absence of an effect with confidence.

Along the way, we evaluated if demographic variables predicted trustworthiness. We only found that nonwhite participants were slightly more likely to trust the answers. We also found that males, individuals with liberal views, and people in urban areas were significantly more likely to check the citations (\ie, mouse-hover over the references) given in the chatbot answer. Given our small sample size, we are hesitant to read too much into these results, but it may be an area for future research. 

\paragraph{Limitations.} Our study is not without several important limitations. First, our sample was comprised of participants entirely from the online data collection platform Prolific. These participants may be more technologically savvy than the typical individual who does not sign up for online research surveys. Additionally, since our sample was 65\% white, we did not have sufficient statistical power to evaluate different racial groups and combined them into a simple nonwhite category. While this nonwhite category trusted their chatbot answers more than the white category, future research will have to investigate what could have caused this difference or if it was an artifact. Our study also did not control for what questions were asked so it could be that people of different demographic groups may ask the chatbot different questions, which could alter their trustworthiness. Finally, our measure of trustworthiness was a simple, one-item variable. It's possible that different forms of trustworthiness may yield interesting results. Future research can incorporate more robust measures of trustworthiness to assess how different questions and outputs influence different elements of trust.

\clearpage

\bibliography{references}

\begin{thebibliography}{51}
\providecommand{\natexlab}[1]{#1}

\bibitem[{Achiam et~al.(2023)Achiam, Adler, Agarwal, Ahmad, Akkaya, Aleman, Almeida, Altenschmidt, Altman, Anadkat et~al.}]{achiam2023gpt4}
Achiam, J.; Adler, S.; Agarwal, S.; Ahmad, L.; Akkaya, I.; Aleman, F.~L.; Almeida, D.; Altenschmidt, J.; Altman, S.; Anadkat, S.; et~al. 2023.
\newblock Gpt-4 technical report.
\newblock \emph{arXiv preprint arXiv:2303.08774}.

\bibitem[{Amaro et~al.(2023)Amaro, Della~Greca, Francese, Tortora, and Tucci}]{amaro2023ai}
Amaro, I.; Della~Greca, A.; Francese, R.; Tortora, G.; and Tucci, C. 2023.
\newblock AI unreliable answers: A case study on ChatGPT.
\newblock In \emph{International Conference on Human-Computer Interaction}, 23--40. Springer.

\bibitem[{Aoki(2020)}]{aoki2020experimental}
Aoki, N. 2020.
\newblock An experimental study of public trust in AI chatbots in the public sector.
\newblock \emph{Government information quarterly}, 37(4): 101490.

\bibitem[{Archard et~al.(2013)Archard, Deveaux, Manson, and Weinstock}]{archard2013reading}
Archard, D.; Deveaux, M.; Manson, N.~A.; and Weinstock, D.~M. 2013.
\newblock \emph{Reading Onora O'Neill}.
\newblock Routledge London/New York, NY.

\bibitem[{Asai et~al.(2023)Asai, Wu, Wang, Sil, and Hajishirzi}]{asai2023self}
Asai, A.; Wu, Z.; Wang, Y.; Sil, A.; and Hajishirzi, H. 2023.
\newblock Self-rag: Learning to retrieve, generate, and critique through self-reflection.
\newblock \emph{arXiv preprint arXiv:2310.11511}.

\bibitem[{Asai et~al.(2024)Asai, Zhong, Chen, Koh, Zettlemoyer, Hajishirzi, and Yih}]{asai2024reliable}
Asai, A.; Zhong, Z.; Chen, D.; Koh, P.~W.; Zettlemoyer, L.; Hajishirzi, H.; and Yih, W.-t. 2024.
\newblock Reliable, adaptable, and attributable language models with retrieval.
\newblock \emph{arXiv preprint arXiv:2403.03187}.

\bibitem[{Azaria and Mitchell(2023)}]{azaria2023internal}
Azaria, A.; and Mitchell, T. 2023.
\newblock The internal state of an llm knows when its lying.
\newblock \emph{arXiv preprint arXiv:2304.13734}.

\bibitem[{Baier(1986)}]{baier1986trust}
Baier, A. 1986.
\newblock Trust and Antitrust.
\newblock \emph{Ethics}, 96(2): 231--260.

\bibitem[{Bonferroni(1936)}]{bonferroni1936teoria}
Bonferroni, C. 1936.
\newblock Teoria statistica delle classi e calcolo delle probabilita.
\newblock \emph{Pubblicazioni del R istituto superiore di scienze economiche e commericiali di firenze}, 8: 3--62.

\bibitem[{Castelo and Ward(2021)}]{castelo2021conservatism}
Castelo, N.; and Ward, A.~F. 2021.
\newblock Conservatism predicts aversion to consequential Artificial Intelligence.
\newblock \emph{Plos one}, 16(12): e0261467.

\bibitem[{Cheng et~al.(2019)Cheng, Wang, Zhang, O'connell, Gray, Harper, and Zhu}]{cheng2019explaining}
Cheng, H.-F.; Wang, R.; Zhang, Z.; O'connell, F.; Gray, T.; Harper, F.~M.; and Zhu, H. 2019.
\newblock Explaining decision-making algorithms through UI: Strategies to help non-expert stakeholders.
\newblock In \emph{Proceedings of the 2019 chi conference on human factors in computing systems}, 1--12.

\bibitem[{Choudhury and Shamszare(2023)}]{choudhury2023investigating}
Choudhury, A.; and Shamszare, H. 2023.
\newblock Investigating the impact of user trust on the adoption and use of ChatGPT: Survey analysis.
\newblock \emph{Journal of Medical Internet Research}, 25: e47184.

\bibitem[{Cialdini(2009)}]{cialdini2009social}
Cialdini, R. 2009.
\newblock Social proof: Truths are us.
\newblock \emph{Influence: Science and practice}, 97--140.

\bibitem[{Claudy, Parkinson, and Aquino(2024)}]{claudy2024should}
Claudy, M.~C.; Parkinson, M.; and Aquino, K. 2024.
\newblock Why should innovators care about morality? Political ideology, moral foundations, and the acceptance of technological innovations.
\newblock \emph{Technological Forecasting and Social Change}, 203: 123384.

\bibitem[{Ditto et~al.(2019)Ditto, Liu, Clark, Wojcik, Chen, Grady, Celniker, and Zinger}]{ditto2019least}
Ditto, P.~H.; Liu, B.~S.; Clark, C.~J.; Wojcik, S.~P.; Chen, E.~E.; Grady, R.~H.; Celniker, J.~B.; and Zinger, J.~F. 2019.
\newblock At least bias is bipartisan: A meta-analytic comparison of partisan bias in liberals and conservatives.
\newblock \emph{Perspectives on Psychological Science}, 14(2): 273--291.

\bibitem[{Ehsan et~al.(2019)Ehsan, Tambwekar, Chan, Harrison, and Riedl}]{ehsan2019automated}
Ehsan, U.; Tambwekar, P.; Chan, L.; Harrison, B.; and Riedl, M.~O. 2019.
\newblock Automated rationale generation: a technique for explainable AI and its effects on human perceptions.
\newblock In \emph{Proceedings of the 24th international conference on intelligent user interfaces}, 263--274.

\bibitem[{Ferrario and Loi(2022)}]{ferrario_how_2022}
Ferrario, A.; and Loi, M. 2022.
\newblock How {Explainability} {Contributes} to {Trust} in {AI}.
\newblock In \emph{2022 {ACM} {Conference} on {Fairness}, {Accountability}, and {Transparency}}, 1457--1466. Seoul Republic of Korea: ACM.
\newblock ISBN 978-1-4503-9352-2.

\bibitem[{Gonen et~al.(2022)Gonen, Iyer, Blevins, Smith, and Zettlemoyer}]{gonen2022demystifying}
Gonen, H.; Iyer, S.; Blevins, T.; Smith, N.~A.; and Zettlemoyer, L. 2022.
\newblock Demystifying prompts in language models via perplexity estimation.
\newblock \emph{arXiv preprint arXiv:2212.04037}.

\bibitem[{Hoff and Bashir(2015)}]{hoff2015trust}
Hoff, K.~A.; and Bashir, M. 2015.
\newblock Trust in automation: Integrating empirical evidence on factors that influence trust.
\newblock \emph{Human factors}, 57(3): 407--434.

\bibitem[{Kaplan et~al.(2023)Kaplan, Kessler, Brill, and Hancock}]{kaplan2023trust}
Kaplan, A.~D.; Kessler, T.~T.; Brill, J.~C.; and Hancock, P.~A. 2023.
\newblock Trust in artificial intelligence: Meta-analytic findings.
\newblock \emph{Human factors}, 65(2): 337--359.

\bibitem[{Kim, Choi, and Fotso(2024)}]{kim2024medical}
Kim, Y.~J.; Choi, J.~H.; and Fotso, G. M.~N. 2024.
\newblock Medical professionals' adoption of AI-based medical devices: UTAUT model with trust mediation.
\newblock \emph{Journal of Open Innovation: Technology, Market, and Complexity}, 10(1): 100220.

\bibitem[{Lewis et~al.(2020)Lewis, Perez, Piktus, Petroni, Karpukhin, Goyal, K{\"u}ttler, Lewis, Yih, Rockt{\"a}schel et~al.}]{lewis2020retrieval}
Lewis, P.; Perez, E.; Piktus, A.; Petroni, F.; Karpukhin, V.; Goyal, N.; K{\"u}ttler, H.; Lewis, M.; Yih, W.-t.; Rockt{\"a}schel, T.; et~al. 2020.
\newblock Retrieval-augmented generation for knowledge-intensive nlp tasks.
\newblock \emph{Advances in Neural Information Processing Systems}, 33: 9459--9474.

\bibitem[{Lins and Sunyaev(2023)}]{lins2023advancing}
Lins, S.; and Sunyaev, A. 2023.
\newblock Advancing the presentation of IS certifications: theory-driven guidelines for designing peripheral cues to increase users’ trust perceptions.
\newblock \emph{Behaviour \& Information Technology}, 42(13): 2255--2278.

\bibitem[{Lipton(2018)}]{lipton2018mythos}
Lipton, Z.~C. 2018.
\newblock The mythos of model interpretability: In machine learning, the concept of interpretability is both important and slippery.
\newblock \emph{Queue}, 16(3): 31--57.

\bibitem[{Lucassen and Schraagen(2011)}]{lucassen2011factual}
Lucassen, T.; and Schraagen, J.~M. 2011.
\newblock Factual accuracy and trust in information: The role of expertise.
\newblock \emph{Journal of the American Society for Information Science and Technology}, 62(7): 1232--1242.

\bibitem[{McInnes, Healy, and Melville(2018)}]{mcinnes2018umap}
McInnes, L.; Healy, J.; and Melville, J. 2018.
\newblock Umap: Uniform manifold approximation and projection for dimension reduction.
\newblock \emph{arXiv preprint arXiv:1802.03426}.

\bibitem[{Miller et~al.(2016)Miller, Johns, Mok, Gowda, Sirkin, Lee, and Ju}]{miller2016behavioral}
Miller, D.; Johns, M.; Mok, B.; Gowda, N.; Sirkin, D.; Lee, K.; and Ju, W. 2016.
\newblock Behavioral measurement of trust in automation: the trust fall.
\newblock In \emph{Proceedings of the human factors and ergonomics society annual meeting}, volume~60, 1849--1853. SAGE Publications Sage CA: Los Angeles, CA.

\bibitem[{Miller(2019)}]{miller2019explanation}
Miller, T. 2019.
\newblock Explanation in artificial intelligence: Insights from the social sciences.
\newblock \emph{Artificial intelligence}, 267: 1--38.

\bibitem[{Mothilal, Sharma, and Tan(2020)}]{mothilal2020explaining}
Mothilal, R.~K.; Sharma, A.; and Tan, C. 2020.
\newblock Explaining machine learning classifiers through diverse counterfactual explanations.
\newblock In \emph{Proceedings of the 2020 conference on fairness, accountability, and transparency}, 607--617.

\bibitem[{Muir(1987)}]{muir1987trust}
Muir, B.~M. 1987.
\newblock Trust between humans and machines, and the design of decision aids.
\newblock \emph{International journal of man-machine studies}, 27(5-6): 527--539.

\bibitem[{Nothdurft, Heinroth, and Minker(2013)}]{nothdurft2013impact}
Nothdurft, F.; Heinroth, T.; and Minker, W. 2013.
\newblock The impact of explanation dialogues on human-computer trust.
\newblock In \emph{Human-Computer Interaction. Users and Contexts of Use: 15th International Conference, HCI International 2013, Las Vegas, NV, USA, July 21-26, 2013, Proceedings, Part III 15}, 59--67. Springer.

\bibitem[{Palan and Schitter(2018)}]{palan2018prolific}
Palan, S.; and Schitter, C. 2018.
\newblock Prolific. ac—A subject pool for online experiments.
\newblock \emph{Journal of Behavioral and Experimental Finance}, 17: 22--27.

\bibitem[{Papenmeier et~al.(2022)Papenmeier, Kern, Englebienne, and Seifert}]{papenmeier2022s}
Papenmeier, A.; Kern, D.; Englebienne, G.; and Seifert, C. 2022.
\newblock It’s complicated: The relationship between user trust, model accuracy and explanations in ai.
\newblock \emph{ACM Transactions on Computer-Human Interaction (TOCHI)}, 29(4): 1--33.

\bibitem[{Poudel and Weninger(2024)}]{poudel2024navigating}
Poudel, A.; and Weninger, T. 2024.
\newblock Navigating the Post-API Dilemma.
\newblock In \emph{Proceedings of the ACM on Web Conference 2024}, 2476--2484.

\bibitem[{Poursabzi-Sangdeh et~al.(2021)Poursabzi-Sangdeh, Goldstein, Hofman, Wortman~Vaughan, and Wallach}]{poursabzi2021manipulating}
Poursabzi-Sangdeh, F.; Goldstein, D.~G.; Hofman, J.~M.; Wortman~Vaughan, J.~W.; and Wallach, H. 2021.
\newblock Manipulating and measuring model interpretability.
\newblock In \emph{Proceedings of the 2021 CHI conference on human factors in computing systems}, 1--52.

\bibitem[{Rainie et~al.(2022)Rainie, Funk, Anderson, and Tyson}]{rainie2022americans}
Rainie, L.; Funk, C.; Anderson, M.; and Tyson, A. 2022.
\newblock How Americans think about artificial intelligence.
\newblock \emph{Pew Research Center: Washington, DC, USA}.

\bibitem[{Rawal et~al.(2021)Rawal, McCoy, Rawat, Sadler, and Amant}]{rawal2021recent}
Rawal, A.; McCoy, J.; Rawat, D.~B.; Sadler, B.~M.; and Amant, R.~S. 2021.
\newblock Recent advances in trustworthy explainable artificial intelligence: Status, challenges, and perspectives.
\newblock \emph{IEEE Transactions on Artificial Intelligence}, 3(6): 852--866.

\bibitem[{Reimers and Gurevych(2019)}]{reimers-2019-sentence-bert}
Reimers, N.; and Gurevych, I. 2019.
\newblock Sentence-BERT: Sentence Embeddings using Siamese BERT-Networks.
\newblock In \emph{Proceedings of the 2019 Conference on Empirical Methods in Natural Language Processing}. Association for Computational Linguistics.

\bibitem[{Rosen(1999)}]{rosen1999journalists}
Rosen, J. 1999.
\newblock \emph{What are journalists for?}
\newblock Yale University Press.

\bibitem[{Sambrook(2012)}]{sambrook2012delivering}
Sambrook, R. 2012.
\newblock \emph{Delivering trust: Impartiality and objectivity in the digital age}.
\newblock Reuters Institute for the Study of Journalism.

\bibitem[{Scharowski et~al.(2023)Scharowski, Perrig, Svab, Opwis, and Br{\"u}hlmann}]{scharowski2023exploring}
Scharowski, N.; Perrig, S.~A.; Svab, M.; Opwis, K.; and Br{\"u}hlmann, F. 2023.
\newblock Exploring the effects of human-centered AI explanations on trust and reliance.
\newblock \emph{Frontiers in Computer Science}, 5: 1151150.

\bibitem[{Srinivas and Friedman(2024)}]{srinivas2024perplexity}
Srinivas, A.; and Friedman, L. 2024.
\newblock Aravind Srinivas: Perplexity CEO on Future of AI, Search \& the Internet Lex Fridman Podcast \#434.
\newblock \url{https://www.youtube.com/watch?v=e-gwvmhyU7A}.

\bibitem[{Stephanidis et~al.(2019)Stephanidis, Salvendy, Antona, Chen, Dong, Duffy, Fang, Fidopiastis, Fragomeni, Fu et~al.}]{stephanidis2019seven}
Stephanidis, C.; Salvendy, G.; Antona, M.; Chen, J.~Y.; Dong, J.; Duffy, V.~G.; Fang, X.; Fidopiastis, C.; Fragomeni, G.; Fu, L.~P.; et~al. 2019.
\newblock Seven HCI grand challenges.
\newblock \emph{International Journal of Human--Computer Interaction}, 35(14): 1229--1269.

\bibitem[{Sun et~al.(2022)Sun, Wang, Tay, Yang, and Zhou}]{sun2022recitation}
Sun, Z.; Wang, X.; Tay, Y.; Yang, Y.; and Zhou, D. 2022.
\newblock Recitation-augmented language models.
\newblock \emph{arXiv preprint arXiv:2210.01296}.

\bibitem[{Thornley et~al.(2015)Thornley, Watkinson, Nicholas, Volentine, Jamali, Herman, Allard, Levine, and Tenopir}]{thornley2015role}
Thornley, C.; Watkinson, A.; Nicholas, D.; Volentine, R.; Jamali, H.~R.; Herman, E.; Allard, S.; Levine, K.; and Tenopir, C. 2015.
\newblock The role of trust and authority in the citation behaviour of researchers.
\newblock \emph{Information research}.

\bibitem[{Touvron et~al.(2023)Touvron, Martin, Stone, Albert, Almahairi, Babaei, Bashlykov, Batra, Bhargava, Bhosale et~al.}]{touvron2023llama2}
Touvron, H.; Martin, L.; Stone, K.; Albert, P.; Almahairi, A.; Babaei, Y.; Bashlykov, N.; Batra, S.; Bhargava, P.; Bhosale, S.; et~al. 2023.
\newblock Llama 2: Open foundation and fine-tuned chat models.
\newblock \emph{arXiv preprint arXiv:2307.09288}.

\bibitem[{Tyson and Kikuchi(2023)}]{tyson2023growing}
Tyson, A.; and Kikuchi, E. 2023.
\newblock Growing public concern about the role of artificial intelligence in daily life.
\newblock Technical report.

\bibitem[{Venkatesh and Davis(2000)}]{venkatesh2000theoretical}
Venkatesh, V.; and Davis, F.~D. 2000.
\newblock A theoretical extension of the technology acceptance model: Four longitudinal field studies.
\newblock \emph{Management science}, 46(2): 186--204.

\bibitem[{Wachter, Mittelstadt, and Russell(2017)}]{wachter2017counterfactual}
Wachter, S.; Mittelstadt, B.; and Russell, C. 2017.
\newblock Counterfactual explanations without opening the black box: Automated decisions and the GDPR.
\newblock \emph{Harv. JL \& Tech.}, 31: 841.

\bibitem[{Zhang, Liao, and Bellamy(2020)}]{zhang2020effect}
Zhang, Y.; Liao, Q.~V.; and Bellamy, R.~K. 2020.
\newblock Effect of confidence and explanation on accuracy and trust calibration in AI-assisted decision making.
\newblock In \emph{Proceedings of the 2020 conference on fairness, accountability, and transparency}, 295--305.

\bibitem[{Zhao et~al.(2024)Zhao, Chen, Yang, Liu, Deng, Cai, Wang, Yin, and Du}]{zhao2024explainability}
Zhao, H.; Chen, H.; Yang, F.; Liu, N.; Deng, H.; Cai, H.; Wang, S.; Yin, D.; and Du, M. 2024.
\newblock Explainability for large language models: A survey.
\newblock \emph{ACM Transactions on Intelligent Systems and Technology}, 15(2): 1--38.

\end{thebibliography}

\newpage

\renewcommand{\thetable}{S\arabic{table}}

\onecolumn
\begin{center}
\Large \textbf{Citations and Trust in AI-Generated Answers}

\large \textbf{Supplemental Information}
\end{center}

\section{Demographic Battery and Response Coding}
\label{app:demo}

Responses from the survey are coded for analysis. Table~\ref{tab:demograph} has the code for each response indicated in parenthesis next to each option. We collected data for man, woman, and other gender, but combined women and other gender into one category. We also constructed racial categories into white and nonwhite given the lack of racial diversity among our participants.   

Trust is numerical and measured on a sliding scale from 1 to 10 where 1 is least trusting and 10 is most trusting. Participants must move the slider from its default position (5) in order to proceed.

\begin{table*}[ht]
    \centering
    \sffamily\scriptsize
    \caption{Demographic Battery and Coding. These questions were asked at the conclusion of the survey. Response codes are indicated in parenthesis. Response frequency is indicated below each option; NA indicates Not Answered. }
    \label{tab:demograph} 
    \begin{tabular}{l|llllllll} 
    \toprule
    \textbf{Question} & \multicolumn{5}{c}{\textbf{Response Choices and (Coding)}}\\
    \midrule
    \multirow{2}{3.2cm}{What gender do you identify with?} & Woman (0) & Man (1) & Other (0) & & &  & NA \\ 
     & 177 & 118 & 7 & & & & 1 \\ \midrule
    \multirow{2}{3.2cm}{How many years old are you?} & 18-30 (1) & 31-40 (2) & 41-50 (3) & 51-60 (4) & 61-70 (5) & >70 (6) & NA \\
    & 104 &93 & 49 & 31& 18&8 &  0 \\\midrule
    \multirow{3}{3.2cm}{Which of the following best describes the area you live in?} & Urban (3) & Suburban (2) & Rural (1) & & & &  NA \\
    & 111 &155 & 35 & & & & 2 \\ 
    \\ \midrule
    \multirow{3}{3.2cm}{What is your highest completed level of education?} & High Sch. (1) & Some Coll. (2) & College (3) & Postgrad (4) & & & NA \\
    & 32 &64 & 147 &59 & & & 1 \\ 
    \\ \midrule
    \multirow{2}{3.2cm}{What race do you identify as?} & Asian (0) & Black (0) & Latino (0) & White (1) & Other (0) & & NA\\
    & 33 &50 & 13 & 195&8& & 0 \\\midrule
    \multirow{2}{3.2cm}{Please indicate your political orientation.} & Vry Lib (1)  & Swht Lib (2) & Mod (3) & Swht Con (4) & Vry Con (5) & &  NA  \\
    & 75 &94 & 68 &39 &20 & & 7 \\\midrule
    \multirow{2}{3.2cm}{Have you heard of ChatGPT?} & Never (1) & Few Times (2) & Familiar (3) & & & & NA  \\
    & 2 & 44 & 257 & & & & 0 \\
    \bottomrule
    \end{tabular}    
\end{table*}

\clearpage

\section{Citations and Trust}

Results of the linear regression test are described in~Tab.~\ref{tab:citation_lr_supp}. The Stata code to generate these results are:

\begin{Verbatim}[breaklines=true]
reg rating citation15 random_citation age chatgpt_heard educ male conserve nonwhite urban    
\end{Verbatim}

\begin{table}[h!]
\centering\sffamily\small
\begin{threeparttable}[h]
\caption{Citations increase perceived trustworthiness, but random citations decrease perceived trustworthiness
}
\label{tab:citation_lr_supp}
\begin{tabular}{r ll}
\toprule
& \multicolumn{2}{c}{\textbf{Trust}}\\ 
& {$\mathbf{\beta}$} & \textbf{Std. Err.} \\ \midrule
Has Citation &  0.394*** & (0.0906) \\ \midrule
Citation Random & -0.268*** & (0.0872) \\ \midrule
Age Group &  -0.017 & (0.0335) \\
Heard of ChatGPT & 0.112 & (0.1240) \\
Education Level & -0.041 & (0.0511) \\
Male &  0.082 & (0.0906) \\
Conservative & -0.061 & (0.0386) \\
Nonwhite &  0.232** & (0.0940) \\
Urban & -0.132* & (0.0698) \\\midrule
Constant & 8.250*** & (0.2550) \\ \midrule
\multicolumn{3}{l}{*** p<0.01, ** p<0.05, * p<0.1} \\ \bottomrule
\end{tabular}
\end{threeparttable}
\end{table}

\clearpage

The hypothesis that five citations would be more trusted than one or zero citations. ANOVA test was used to determine the this relationship. Shapiro-Wilk test was showed that the trust ratings departed significantly from normality ($W$=[.84, .80, .80], $p$<0.0001) for zero, one, and five citation-conditions respectively. Based on this outcome, we conducted Welch's ANOVA to account for unequal variances. We found statistical differences between these citation conditions ($F(2, 3037)=10.23$, $p$<0.0001)

Results of one way ANOVA test for are in~Tab.~\ref{tab:numcite_supp}. The Stata code to generate these results are:

\begin{Verbatim}[breaklines=true]
oneway rating num_citation, bonferroni tabulate
\end{Verbatim}

\begin{table}[h!]
\centering\sffamily\small
\begin{threeparttable}[h]
\centering
\caption{One-way ANOVA Test. Five citations are not perceived as more trustworthy than one citation.
}
\label{tab:numcite_supp}
\begin{tabular}{r | lll}
\toprule
& \multicolumn{3}{c}{\textbf{Trust}} \\
\textbf{Num Citations} & \textbf{Mean} & \textbf{Std. Err.} & \textbf{N} \\ \midrule
 0 &  7.732 & (2.463) & 1064 \\ 
 1 &  8.103 & (2.256) & 959\\ 
 5 &  8.170 & (2.217) & 1017\\\bottomrule
\end{tabular}
\vspace{.5cm}
\begin{tabular}{r | llll}
\toprule
 & \multicolumn{4}{c}{\textbf{Bonferroni}} \\
 \textbf{Row-Col} & \multicolumn{2}{c}{\textbf{0}} &  \multicolumn{2}{c}{\textbf{1}}   \\ \midrule
 1 &  0.371 & (0.001) &  & \\ 
 5 &  0.438 & (0.000) & 0.067 & (1.000)\\ \bottomrule
\end{tabular}
\end{threeparttable}
\end{table}

\clearpage

\section{Citation Checking and Trust}

Results of the linear regression test described in~Tab.~\ref{tab:anova_supp}. The Stata code to generate these results are:

\begin{Verbatim}[breaklines=true]
reg mouseover random_citation citation15 rating age chatgpt_heard educ male conserve nonwhite urban
\end{Verbatim}

\begin{table}[h]
\centering\sffamily\small
\begin{threeparttable}[t]
\caption{More Checking of Citations predicts less trust and is more likely to be done by men, liberals, and urban participants}
\label{tab:anova_supp}
\begin{tabular}{r | ll}
\toprule
& \multicolumn{2}{c}{\textbf{Citation Checks}}\\ 
& {$\mathbf{\beta}$} & \textbf{Std. Err.} \\ \midrule
Trust& -0.058***&(0.0103)\\ \midrule
Citation Random &  -0.2107 ***& (0.049)\\
Has Citation& 0.423***& (0.051)\\
Age Group &  0.005& (0.0188)\\
Heard of ChatGPT & 0.057& (0.0696)\\
Education Level & -0.0292& (0.028)\\
Male &  0.202***& (0.0508)\\
Conservative & -0.085***& (0.0217)\\
Nonwhite &  -0.071& (0.052)\\
Urban & 0.153***& (0.039)\\\midrule
Constant & 0.367**& (0.1668)\\ \midrule
\multicolumn{3}{l}{*** p<0.01, ** p<0.05, * p<0.1} \\ \bottomrule
\end{tabular}
\end{threeparttable}
\end{table}

\subsection*{Citation Checks and Number of Citations}

In an exploratory analysis, we investigated the likelihood that a participant checked a citation compared to the number of citations present. There were equal number of experiments with 1 and 5 citations present.

\begin{table}[h]
\centering\sffamily\small
\begin{threeparttable}[t]
\caption{Contingency table for number of citation checks and number of citations present}
\label{tab:checkcont_supp}
\begin{tabular}{rl l l}
\toprule
& & \textbf{Check} & \textbf{No Check} \\ \midrule
\multirow{2}{*}{\textbf{Citations}}& \textbf{1} & 24 & 72 \\
& \textbf{5} & 59 & 42 \\\bottomrule
\end{tabular}
\end{threeparttable}
\end{table}

We performed a $\chi^2$ test to compare this relationship in the contingency table in Tab.~\ref{tab:checkcont_supp}. The relation between these variables was significant, $\chi^2$(1, $N$=197)=21.19, $p$<0.0001. Five citations are more likely to be checked than one.

\clearpage

\subsection*{Question Order, Citation Checking, and Trust}

\begin{figure}[h]
    \centering   
\begin{tikzpicture}
    \begin{axis}[
        width=9cm,
        height=5cm,
        xlabel=Questions Number,
        ylabel=Citation Checks,
        xmin=0.5, xmax=10.5,
        ymin=0, ymax=32,
        xtick={1,2,3,4,5,6,7,8,9,10},
        ytick={0,10,20,30}
        ]
    \addplot[color=black,mark=x]
        plot coordinates {
            (1,21)
            (2,23)
            (3,15)
            (4,16)
            (5,22)
            (6,18)
            (7,18)
            (8,22)            
            (9,16)            
            (10,21)
        };
    \end{axis}
    \end{tikzpicture}
\caption{Number of Citation Checks per Question. The frequency that a citation is checked is not correlated with the question number.}
    \label{fig:citcheck_byq}
\end{figure}
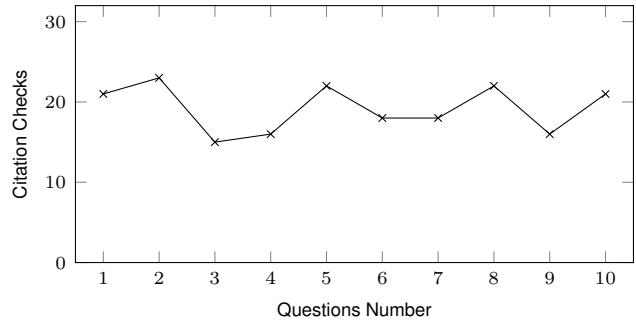

In an exploratory analysis, we investigated the frequency of citation checks across the sequence of questions presented to participants. We hypothesized that citation checks would be more frequent at the beginning of the survey, decreasing as participants progressed, possibly due to a decline in vigilance or increased trust over time. However, as illustrated in Table~\ref{fig:citcheck_byq}, our data did not support this expectation. The pattern of citation checks did not show a significant change up or down, suggesting a different dynamic in participant behavior than anticipated.

\begin{figure}[h]
    \centering
    \input{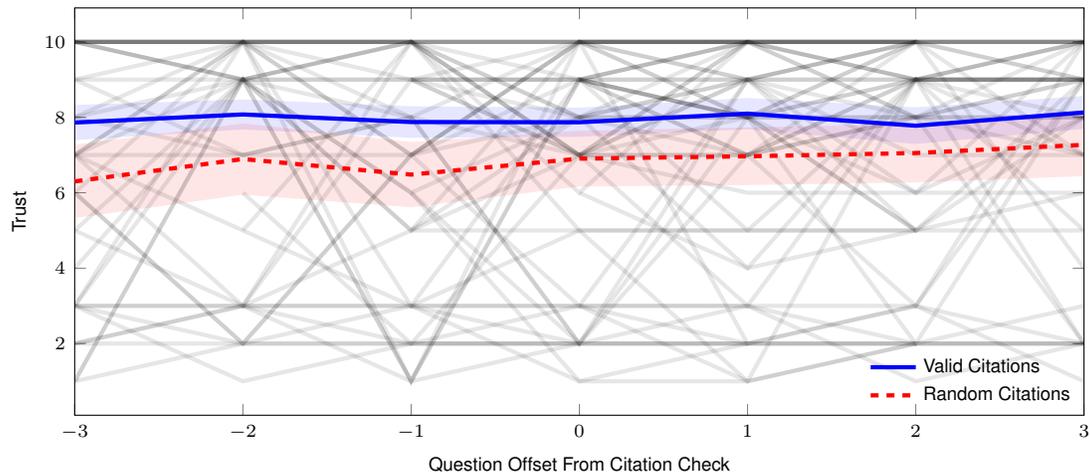}
    \caption{Trust ratings as a function of offset from a citation check. Gray lines represent individual participant trust ratings for all participants who made at least one citation check. Mean and 95\% confidence intervals for valid citations and random citations are in solid-\textcolor{blue}{blue} and dashed-\textcolor{red}{red} lines respectively. We find that, if a user checks a citation in the random condition, they do not to lose trust the chatbot in afterwards.}
    \label{fig:randomchecks}
\end{figure}

Additionally, we investigated whether citation checks influenced trust ratings in subsequent questions. Figure~\ref{fig:randomchecks} presents the trust ratings for questions immediately preceding and following a citation check. For example, if a participant checked the citation for question four, we analyzed the trust ratings for the three preceding and three subsequent questions for that participant. We hypothesized that encountering a random citation would lead to a decline in trust ratings for subsequent questions, as detecting an irrelevant citation might reduce the participant's overall trust in the system. Contrary to our expectations, the analysis did not reveal any significant differences in trust ratings following a citation check.

\clearpage

\section{Trust and Question Type}

Results of the linear regression test are described in~Tab.~\ref{tab:checking_supp}. The Stata code to generate these results are:

\begin{Verbatim}[breaklines=true]
reg rating politics citation15 random_citation age chatgpt_heard educ male conserve nonwhite urban
reg rating fact_or_opinion citation15 random_citation age chatgpt_heard educ male conserve nonwhite urban
reg rating complex_problem_solving citation15 random_citation age chatgpt_heard educ male conserve nonwhite urban
\end{Verbatim}

\begin{table}[h]
\centering\sffamily\small
\begin{threeparttable}[t]
\caption{Political and Fact-based questions are rated as more trustworthy.}
\label{tab:checking_supp}
\begin{tabular}{r | ll| ll| ll}
\toprule
 & \multicolumn{2}{c|}{\textbf{Political Question}} & \multicolumn{2}{c|}{\textbf{Fact Question}} & \multicolumn{2}{c}{\textbf{Complex Question}} \\
 & $\mathbf{\beta}$ & \textbf{Std. Err.} & $\mathbf{\beta}$ & \textbf{Std. Err.} & $\mathbf{\beta}$ & \textbf{Std. Err.}  \\\midrule
 \textbf{Trust} & 0.314**& (0.156) & 0.374**& (0.086) & 0.070& (0.275)\\ \midrule
Has Citation &  0.399***& (0.091) & 0.392***& (0.090) & 0.394***& (0.091)  \\
Citation Random & -0.263***& (0.087) & -0.272***& (0.087) & -0.268***& (0.087)  \\
Age Group &  -0.0210& (0.034) & -0.019& (0.033) & -0.018& (0.034)\\
Heard of ChatGPT & 0.111& (0.124)  & 0.019& (0.124) & 0.112& (0.124) \\
Education Level & -0.0441& (0.051)  & -0.036& (0.051) & -0.041& (0.051) \\
Male &  0.082 & (0.091) & 0.089& (0.090) & 0.083& (0.091)\\
Conservative & -0.061 & (0.039)  & -0.058& (0.039) & -0.061& (0.039) \\
Nonwhite &  0.237**& (0.094) & 0.236**& (0.094) & -0.232**& (0.094)\\
Urban & -0.134*& (0.070)  & -0.133*& (0.070) & -0.132*& (0.070) \\\midrule
Constant & 8.241***& (0.255)  & 8.019***& (0.260) & 8.247***& (0.255) \\ \midrule
\multicolumn{3}{l}{*** p<0.01, ** p<0.05, * p<0.1} \\ \bottomrule
\end{tabular}
\end{threeparttable}
\end{table}

We conducted a Kruskal-Wallis test to assess whether there was a significant difference in citation checks between question types. We included political, fact based, and complex questions along with all other questions in our test. We removed the 38 questions that were coded as both fact-based and complex. We only included data from questions that were in the citation conditions. We did not find that any question type was significantly more likely to have a citation check (H(3)=2.33, $p$=0.508).



\end{document}